\pdfoutput=1
\documentclass[letterpaper, 10 pt, conference]{ieeeconf}
\IEEEoverridecommandlockouts
\usepackage{cite}
\usepackage[figuresleft]{rotating}
\usepackage{amsmath,amssymb,amsfonts}
\usepackage[hidelinks]{hyperref}
\usepackage{bm}
\usepackage{algorithmic}
\usepackage[format=hang]{subfig}
\usepackage{floatrow}
\usepackage{graphicx}
\usepackage{textcomp}
\usepackage{xcolor}
\usepackage{psfrag}
\usepackage{booktabs}
\title{\LARGE \bf
Detecting and Counting Oysters}
\author{Behzad Sadrfaridpour$^{1}$, Yiannis Aloimonos$^{1}$, Miao Yu$^{2}$, Yang Tao$^{3}$, and Donald Webster$^{4}$
\thanks{$^{1}$ Behzad Sadrfaridpour and Yiannis Aloimonos are with the University of Maryland Institute for Advanced Computer Studies, University of Maryland, College Park, MD {\tt\small behzad@umiacs.umd.edu} and {\tt\small yiannis@umiacs.umd.edu}}
\thanks{$^{2}$ Miao Yu is with the Department of Mechanical Engineering and Institute for Systems Research, University of Maryland, College Park, MD {\tt\small mmyu@umd.edu}}
\thanks{$^{3}$ Yang Tao is with the Fischell Department of Bioengineering, University of Maryland, College Park, MD {\tt\small ytao@umd.edu}}
\thanks{$^{4}$ Donald Webster is with the Wye Research and Education Center, University of Maryland, College Park, MD {\tt\small dwebster@umd.edu}}
}
\begin{document}
\maketitle
\thispagestyle{empty}
\pagestyle{empty}
\begin{abstract}
Oysters are an essential species in the Chesapeake Bay living ecosystem. Oysters are filter feeders and considered the vacuum cleaners of the Chesapeake Bay that can considerably improve the Bay's water quality. Many oyster restoration programs have been initiated in the past decades and continued to date. Advancements in robotics and artificial intelligence have opened new opportunities for aquaculture. Drone-like ROVs with high maneuverability are getting more affordable and, if equipped with proper sensory devices, can monitor the oysters. This work presents our efforts for videography of the Chesapeake bay bottom using an ROV, constructing a database of oysters, implementing Mask R-CNN for detecting oysters, and counting their number in a video by tracking them.
\end{abstract}
%
%
%
\section{INTRODUCTION}\label{sec:introduction}
Oysters are an essential species in marine ecosystems. They are filter feeders that clean the surrounding water and create habitat for the other species. However, oysters habitats have undergone significant drops both on national and global levels~\cite{beck2011oyster}. There are ongoing efforts to restore oysters habitats across the United States~\cite{mcfarland2018restoring,baggett2015guidelines,blomberg2018habitat,theuerkauf2019integrating}. A significant challenge is providing adequate and effective means for monitoring their progress. To improve the consistency and extensibility of monitoring of oyster habitat restoration projects, the authors in~\cite{baggett2015guidelines} used an expert group in designing and monitoring oyster reef projects to standardize monitoring metrics, units, and performance criteria for evaluation. They recommended monitoring a set of universal parameters including ``reef areal dimensions, reef height, oyster density, and oyster size-frequency distribution'' and environmental variables including water salinity, temperature, and dissolved oxygen~\cite{baggett2015guidelines}. Other oyster restoration projects also report monitoring these metrics and variables in part~\cite{mcfarland2018restoring,blomberg2018habitat,theuerkauf2019integrating}.

Currently, the measurement of the number of oysters and their physical dimensions for sampling is performed manually and, therefore, is restricted to small amounts, e.g., 100 oysters per sample site~\cite{mcfarland2018restoring}. Moreover, these measurements are not regularly available for small-size aqua/oyster farms. With advancements in robotics and artificial intelligence, there is a promising potential to improve the monitoring process of oyster habitat restoration. Drone-like ROVs (Remotely Operated underwater Vehicles) with high maneuverability are getting more affordable and, with proper sensory devices, can be used for sampling. Moreover, computer vision algorithms can be developed to detect oysters and calculate their physical dimensions and properties. More specifically, convolutional neural networks (CNNs) have been widely used for object detection, and we use them as state-of-the-art for oyster detection. To get the number of oysters in a video, we count the identified oysters. However, to prevent recounting, the detected oysters need to be tracked in consecutive image frames. We use the KCF tracker from OpenCV for tracking the already identified oysters.

Our goal is to step toward monitoring and counting but in a way that would also benefit the oyster harvesting problem. We use an ROV for data collection and implement object detection and instance segmentation algorithms that can be later used for automated harvesting of oysters. However, we do not address oyster picking and harvesting in this paper. Here, we explain how we collected the videos of oysters in the Chesapeake bay and used them for training the oyster detection CNNs. Next, we explain how we tracked the detected oysters to count the oyster in a video. The organization of the rest of the paper is as follows. Section~\ref{sec:related} provides a related work for bottom habitat mapping and common object detection and segmentation algorithms used in precision agriculture. Section~\ref{sec:materials} discusses the object detection and counting algorithm. Section~\ref{sec:result} provides the results of our implementation and experiments.
\section{RELATED WORK}\label{sec:related}
Most of the related work for seafloor bottom mapping in subtidal waters is done using remote sonar sensing techniques followed by in situ sampling or underwater videography for ground-truthing~\cite{ojeda2004spatially}. However, sonar data sets have low resolutions, and obtaining data sets with high-quality data is expensive. Moreover, their process, interpretation, and visualization demand substantial effort. Some works~\cite{grizzle2008bottom} used the georeferenced optical data as the primary mapping tool. A significant problem of visual data is its interpretation~\cite{anderson2008acoustic}. Compared to the physical data, the visual data (i.e., photographs) are limited in terms of measuring the physical parameters (such as size) of samples~\cite{anderson2008acoustic}. However, with advancements in underwater robotics and image processing, the capabilities in obtaining and using optical data have much increased. For instance, recently, an approach for autonomous coral identification and counting~\cite{modasshir2018coral} and underwater 3D semantic mapping~\cite{modasshir2019autonomous} have presented. A similar approach can be adopted for the problem of oyster habitat mapping. 

Generally, works for detection and tracking are categorized as Multi-Object Tracking (MOT), an active research topic in computer vision focusing on identifying and tracking objects of specific classes such as cars, animals, and pedestrians videos~\cite{ciaparrone2020deep}. 
Usually, the objects are animated and have arbitrary trajectories, while here, the objects (oysters) are in-animated, and the camera is moving. Nonetheless, we can still employ MOT practices for our task. We employ tracking-by-detection, which includes three steps~\cite{papakis2020gcnnmatch}: detection of objects, tracking of the previously detected objects, and association of the tracked objects with the detected objects.

Deep learning methods for object detection can be categorized as one-stage and two-stage algorithms. One-stage algorithms such as YOLO~\cite{redmon2016you} SSD~\cite{liu2016ssd} and Fast R-CNN~\cite{girshick2015fast}  solve the object detection using a single network~\cite{redmon2016you} aiming for faster inference computation time while two-stage algorithms such as Faster-RCNN~\cite{ren2015faster} solve the problem in two stages, one for proposing candidate bounding boxes and one for later classifying them as foreground classes and background, aiming for higher inference precision. Mask-RCNN~\cite{he2017mask} is an extension of Faster R-CNN, which also segments the foreground classes using a CNN at the end parallel to the bounding boxing recognition branch. Mask R-CNN demonstrated that it has the highest precision for common datasets~\cite{wu2019detectron2}. In this work, the segmentation helps with generating the size maps of the oysters. Therefore, we used Mask R-CNN~\cite{he2017mask} for detecting oysters.

Various approaches have been applied for tracking and feature extraction~\cite{papakis2020gcnnmatch}. These include deep learning methods, correlation filters (CF), auto-encoders, spatial attention, motion extraction, and feature pyramids. CF-based trackers are among the computationally efficient yet robust trackers~\cite{henriques2014high}. We tried different trackers and chose KCF (Kernelized Correlation Filters) tracker~\cite{henriques2014high} due to its robustness and efficiency.
\begin{figure}[!t]
    \centering
    \includegraphics[width=.99\textwidth]{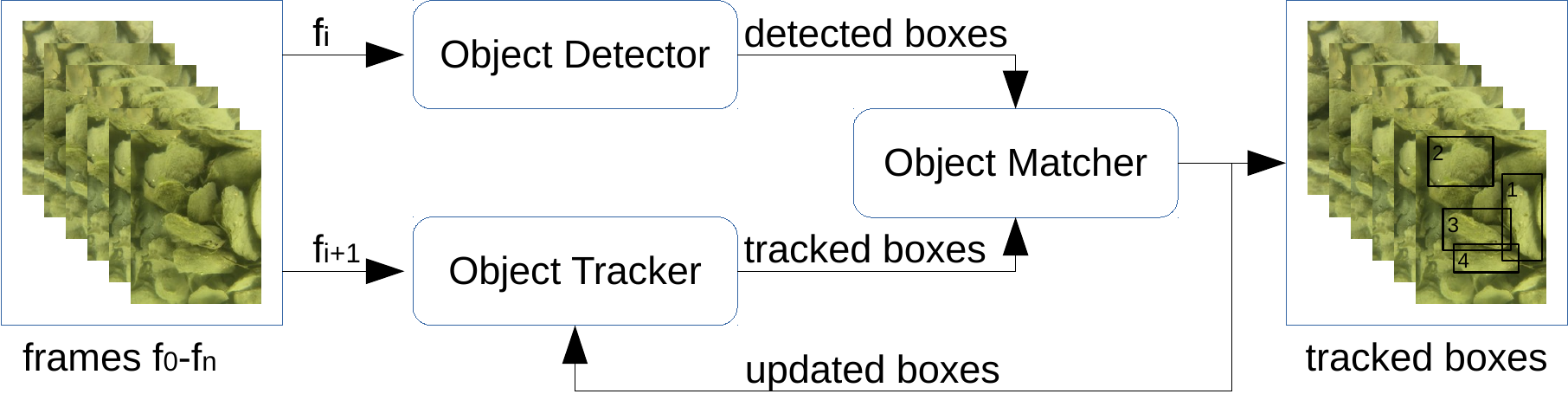}
    \caption{Tracking by detection process}
    \label{fig:process}
\end{figure}
\section{METHODS}\label{sec:materials}
Three main steps required for counting the numbers of oysters in a video are detection, tracking, and association. We used Mask R-CNN for detection, KCF tracker for tracking, and pairwise matching of the detection and tracking bounding boxes based on Intersection over Union (IoU) for the association. 
First, we detect the oysters in the current frame and specify their bounding boxes as described in Section~\ref{sec:network}. Next, we update the tracking bounding boxes from the previous frames described in Section~\ref{sec:tracking}. The final stage is pairwise matching between the newly detected bounding boxes and the updated tracking bounding boxes and associating new trackers for the non-matched oysters (Section~\ref{sec:association}). Figure~\ref{fig:process} provides an overview of these steps.
\subsection{Object Detection Network Structure}\label{sec:network}
\begin{figure}
    \centering
    \includegraphics[width=2.45in]{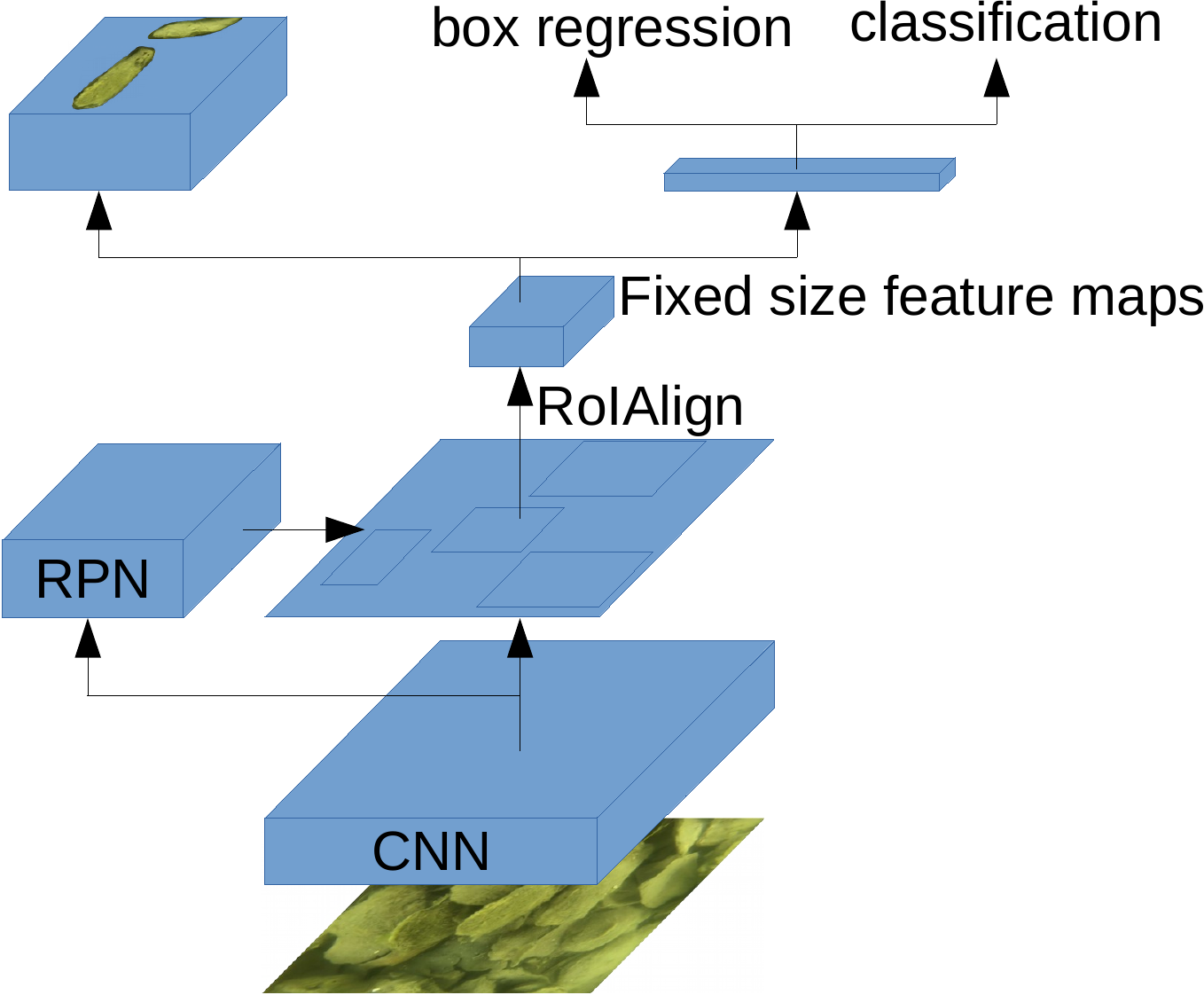}
    \caption{Mask R-CNN network structure}
    \label{fig:network}
\end{figure}

For detecting the oysters, it is most suitable to choose the object detection architectures such as R-CNN (Regions with CNN features)~\cite{girshick2014rich}, Fast R-CNN~\cite{girshick2015fast}, Faster R-CNN~\cite{ren2015faster} and Mask R-CNN~\cite{he2017mask}. The input of these networks is the image, and the outputs are objects and their corresponding bounding boxes or masks of their instance segmentation in the image. R-CNN uses selective search algorithm to generate around 2k Region of Interests (RoIs), wraps them to a fixed square size, and finally runs them through a CNN. CNN is trained for classifying the images and predicting the boxing offset for the original RoIs. Generating the RoIs is computationally expensive and is the bottleneck of the R-CNN method. Fast R-CNN is an improvement of R-CNN. Similar to R-CNN, it calculates the RoIs first using the selective search algorithm. However, it first runs the entire image on a backbone CNN to get a high-resolution feature map corresponding to the whole image. It generates RoIs on the feature map of the image rather than the original image. It then uses pooling to reshape and unify the size of RoIs. The output of Fast R-CNN is similar to R-CNN. Faster R-CNN improves Fast R-CNN by using a separate network, Region Proposal Network (RPN), for proposing the RoIs from the output of the backbone CNN, feature map(s) of the image. It uses RoI pooling to reshape the proposals and then classify and predict the bounding boxes offset similar to Fast R-CNN. Mask R-CNN works similar to Faster R-CNN but extends it with another network for segmenting the RoIs.
Our goal is to be able to generate the size and frequency map of the Chesapeake Bay oysters. Therefore, we used Mask R-CNN that outputs instance segmentation of objects in addition to bounding boxes and classes. Figure~\ref{fig:network} shows the structure of this network.
Different CNNs such as VGGNet, ResNet and with different depths are being used as the backbone of Mask R-CNN. Moreover, Feature Pyramid Networks (FPN)~\cite{lin2017feature} are also being used as an extension to the backbone network. FPN builds a pyramid of feature maps with different sizes by making two series of layers (bottom-up and top-down) with lateral connections. Each feature map then goes through the rest of the network, i.e., RPN, classifier, bounding box regressor, and segmentation. The final segmentation is made by merging the masks outputted for all feature maps. 
The modified versions of ResNets that extracted features from the final convolutional layer of the 4-th stage, called C4, were used in the original Faster-RCNN implementation. Another choice for the backbone network is  ResNet with dilations in 5-th stage~\cite{dai2017deformable}, which is called DC5.
In this work, we used ResNet50-C4, ResNet50-DC5, ResNet50-FPN, ResNet101-FPN, ResNet101-C4, and ResNet101-FPN as the backbone of the Mask R-CNN and compared the results.
\subsection{Tracking}\label{sec:tracking}
The KCF tracker achieves efficient tracking by generating a matrix of cyclic shifts of the sample vector $\bm{x}=[x_1, x_2, \dots,  x_n]^T \in \mathbb{R}^n$. The cyclic shift of $\bm{x}$ is defined as $\bm{P}_x=[x_n, x_1, x_2, \dots,  x_{n-1}]$. Matrix of all cyclic shifts of $\bm{x}$, the circulant matrix, is defined as 
\begin{equation}
\bm{X}=C(\bm{x})=
\begin{bmatrix}
x_1 & x_2 &  \dots & x_n\\
x_n & x_1 &  \dots & x_{n-1}\\
x_{n-1} & x_n  & \dots & x_{n-2}\\
\vdots & \vdots & \ddots & \vdots\\
x_2 & x_3 &  \dots &  x_1
\end{bmatrix}. 
\end{equation}
The KCF tracker uses the cyclic shift method for $(M\times N)$ image block $\bm{x}$ to train a filter $f(\bm{x})=\bm{\omega}\bm{\phi}(\bm{x})$, where $\bm{\phi}(\bm{x})$ is the mapping Fourier space. The training samples for $\bm{x}$ are all of the cyclic shifts $\bm{P}_i$ where $i \in \{0, \dots, M-1\}\times \{0, \dots, N-1\}$. The corresponding score $y_i \in [0, 1]$ for $\bm{P}_i$ is generated by a Gaussian function based on shift distance. The classifier for detection of the tracker is trained by minimizing the regressor error as~\cite{liuoverview}:
\begin{equation}
    w=\operatorname*{argmin}_w \sum_{i}{} (\bm{\omega}\bm{\phi}(\bm{x})-y_i)^2=\lambda \|w\|^2,
\end{equation}
where $\lambda \geq 0$ is the regularization parameter for the model's simplicity. The hypothesis is trained effectively by using fast Fourier transform. 
In the Fourier domain, instead of a single grayscale channel feature input, the KCF algorithm can use multi-channel histogram of oriented gradients (HOG) features as input, simply by summing them over~\cite{henriques2014high}.
\subsection{Association}\label{sec:association}
As shown in Fig.~\ref{fig:process}, for each frame in the video sequence, the detected bounding boxes (of oysters) are compared with the tracked bounding boxes to add new trackers, update trackers or remove trackers. The bounding boxes are matched in pairs using their IoU values. For two bounding boxes, IoU is defined as the ratio of area of their overlap over the area of their union. We define the IoU overlap threshold of 0.2 for matching the boxes. If the IoU of a detected box with a tracking box is greater than the overlap threshold, we match them. Moreover, we add new tracking boxes for the non-matched detected boxes and update the matched tracking boxes. We keep track of any unmatched tracking box and remove it after a certain number of mismatches.
Each tracker is assigned a unique number, and the output frame is the updated bounding boxes. Thus the total number of oysters is the number of trackers assigned for the video sequence.

\section{EXPERIMENTS AND RESULTS}\label{sec:result}
This section describes our underwater image acquisition, implementation and training of the Mask-RCNN network, and, finally, results for detection and tracking.
\subsection{Image Acquisition}
\begin{figure} 
\centering
\subfloat[]{\includegraphics[width=.35\textwidth]{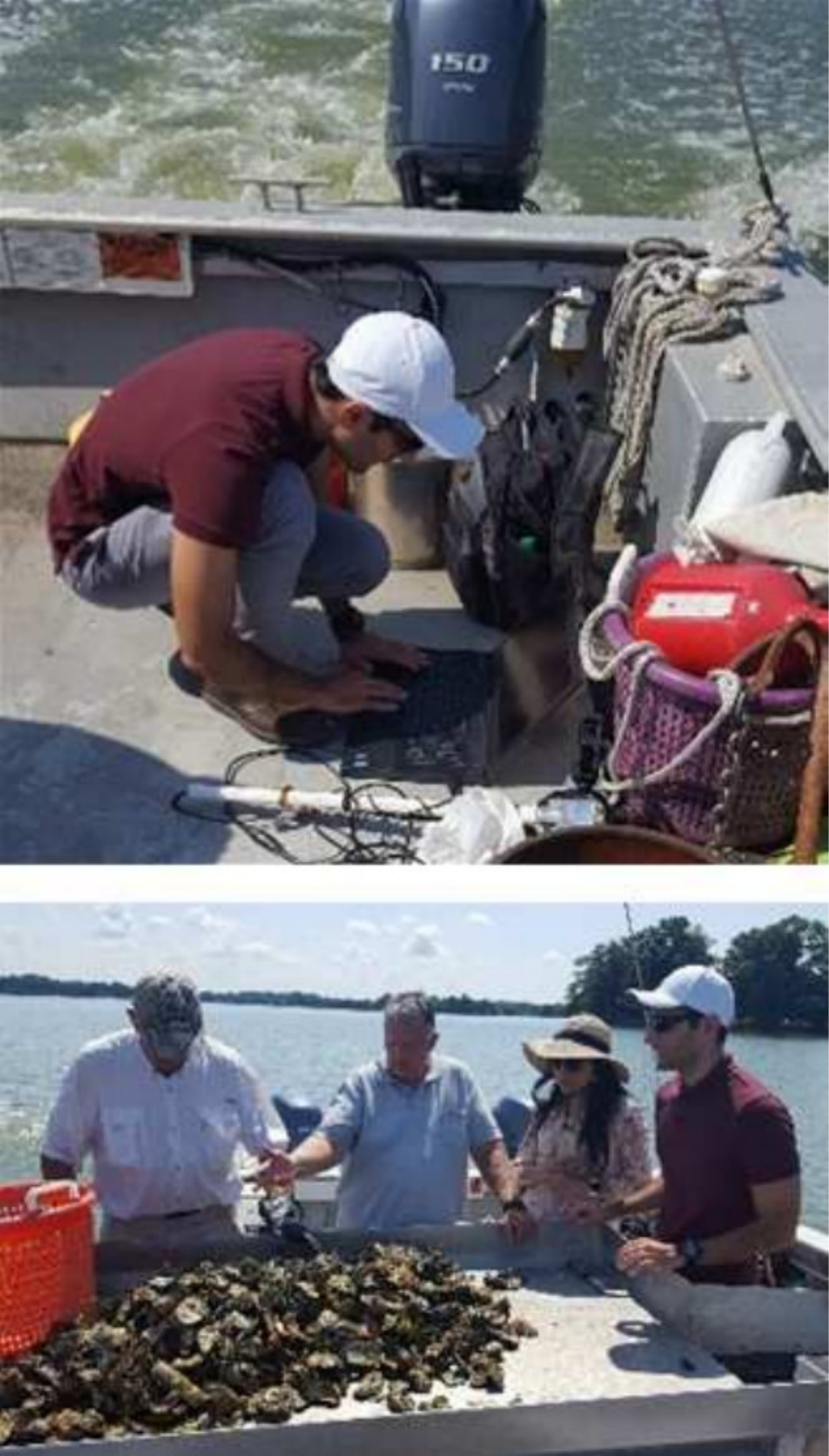}}\hfill
\subfloat[]{\includegraphics[width=.51\textwidth]{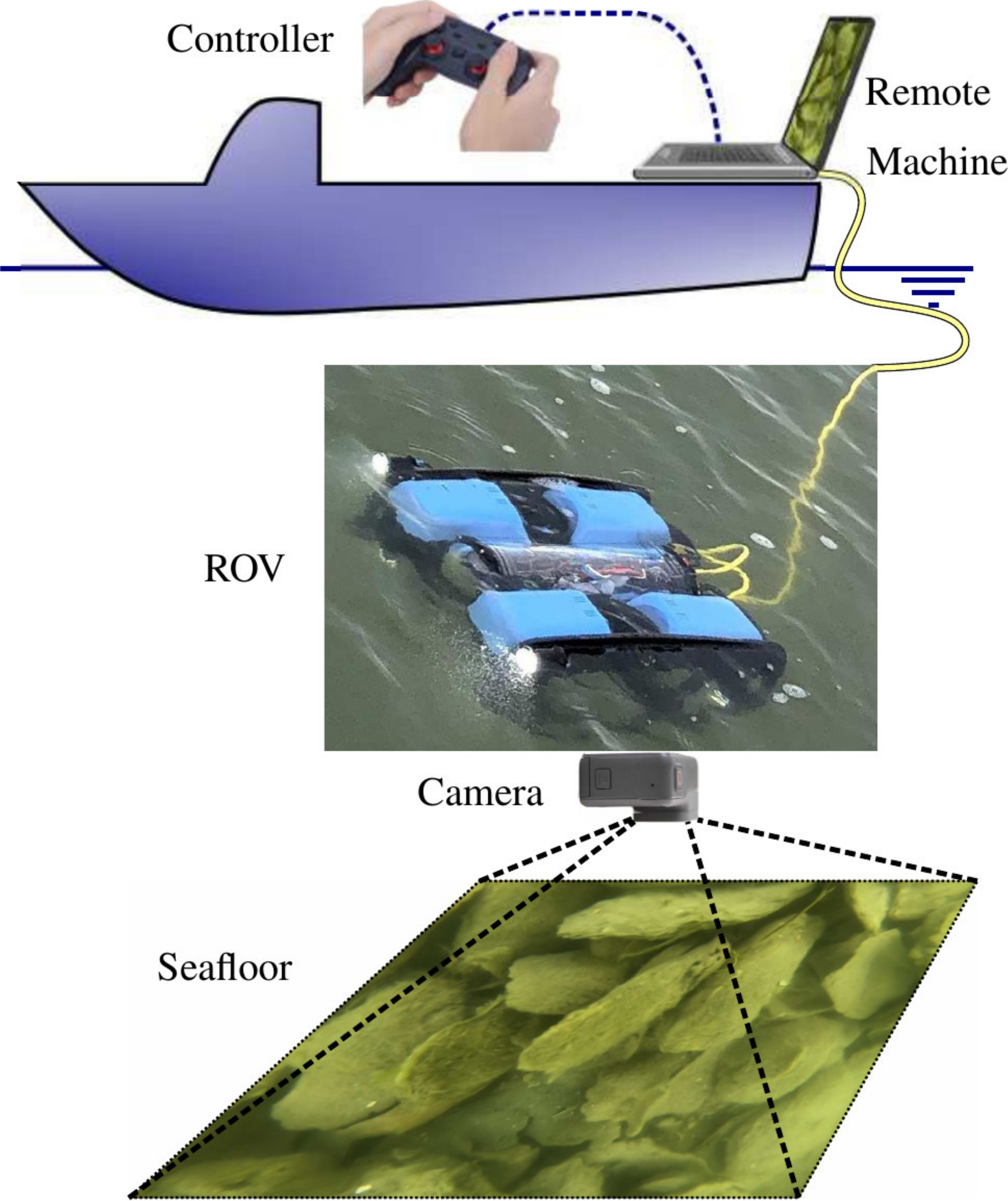}}
\caption{Data acquisition: (a) preliminary (b) by ROV}
\label{fig:daq}
\end{figure}

We conducted a preliminary study on the underwater machine vision of oysters in the Chesapeake Bay, as shown in Fig~\ref{fig:daq}a. We mounted an led light and a waterproof camera on a long pipe and took underwater photos. Underwater illumination is a challenging problem and had a high impact on the quality of the collected video sequence. Next, we used a Remotely Operated underwater Vehicle (ROV) for taking underwater videos. Figure~\ref{fig:daq}(b) shows our underwater videography setup. The ROV is BlueRobotics BlueROV2~\cite{blue}, with two horizontal thrusters and four vertical thrusters providing six degrees of freedom, three translational (heave, sway, and surge) and three rotational (pitch, roll, and yaw). It is connected via a tether to a remote machine located on the ground or boat. An open-source software, QGroundControl (QGC)~\cite{qgc}, is installed on this machine as the graphic user interface (GUI) of ROV for sending commands to ROV and receiving camera and other sensory information on the ground station machine (remote machine). The ROV can be controlled autonomously or manually. Here, we control it manually by sending control commands to QGC using a wireless controller. A camera and two led lights are mounted in front of the ROV. The camera feed streams to the remote machine, and the operator uses it for the underwater navigation of the ROV.

\begin{figure} 
\centering
\subfloat[]{\includegraphics[width=.225\textwidth]{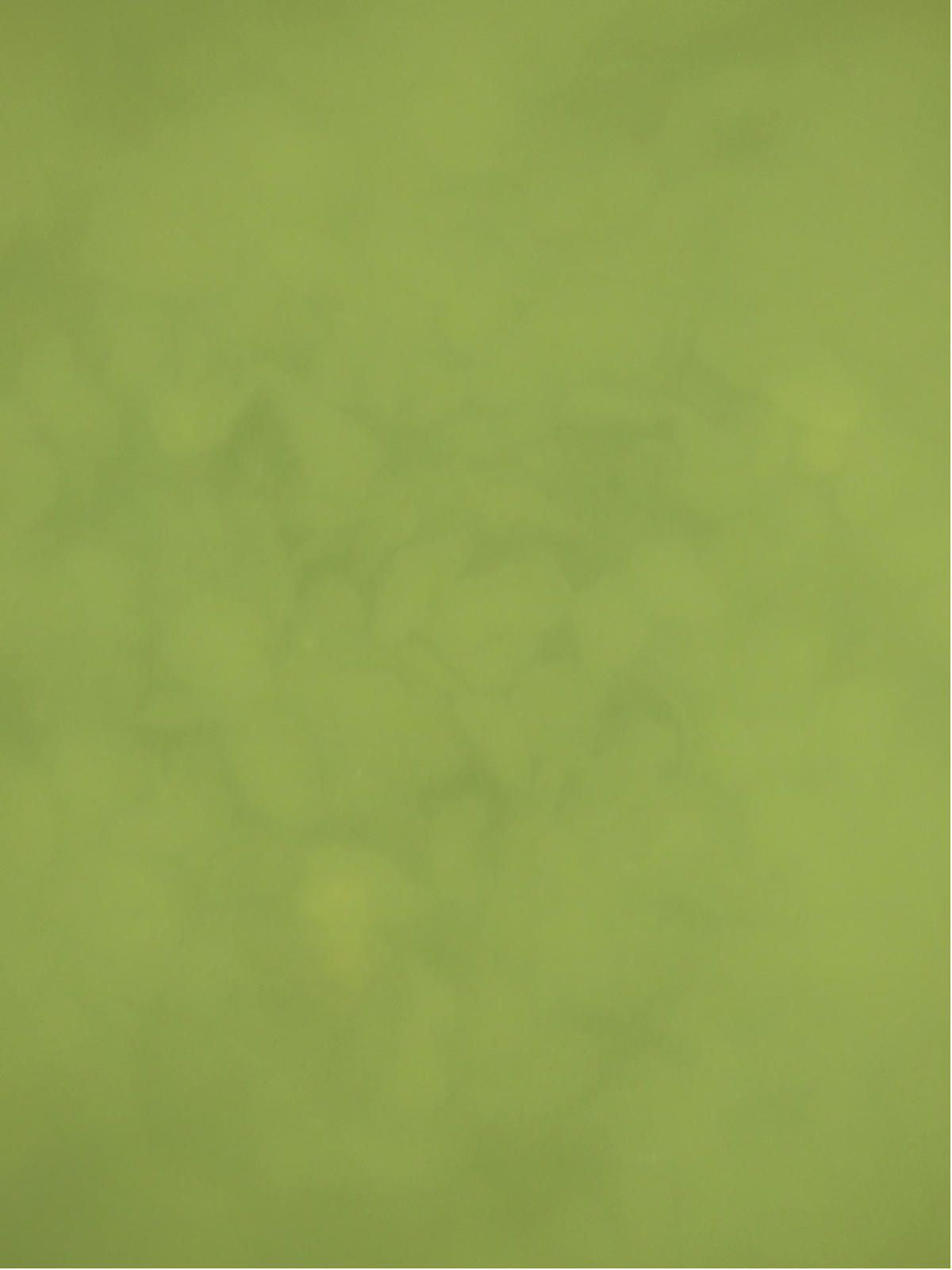}}\hfill
\subfloat[]{\includegraphics[width=.225\textwidth]{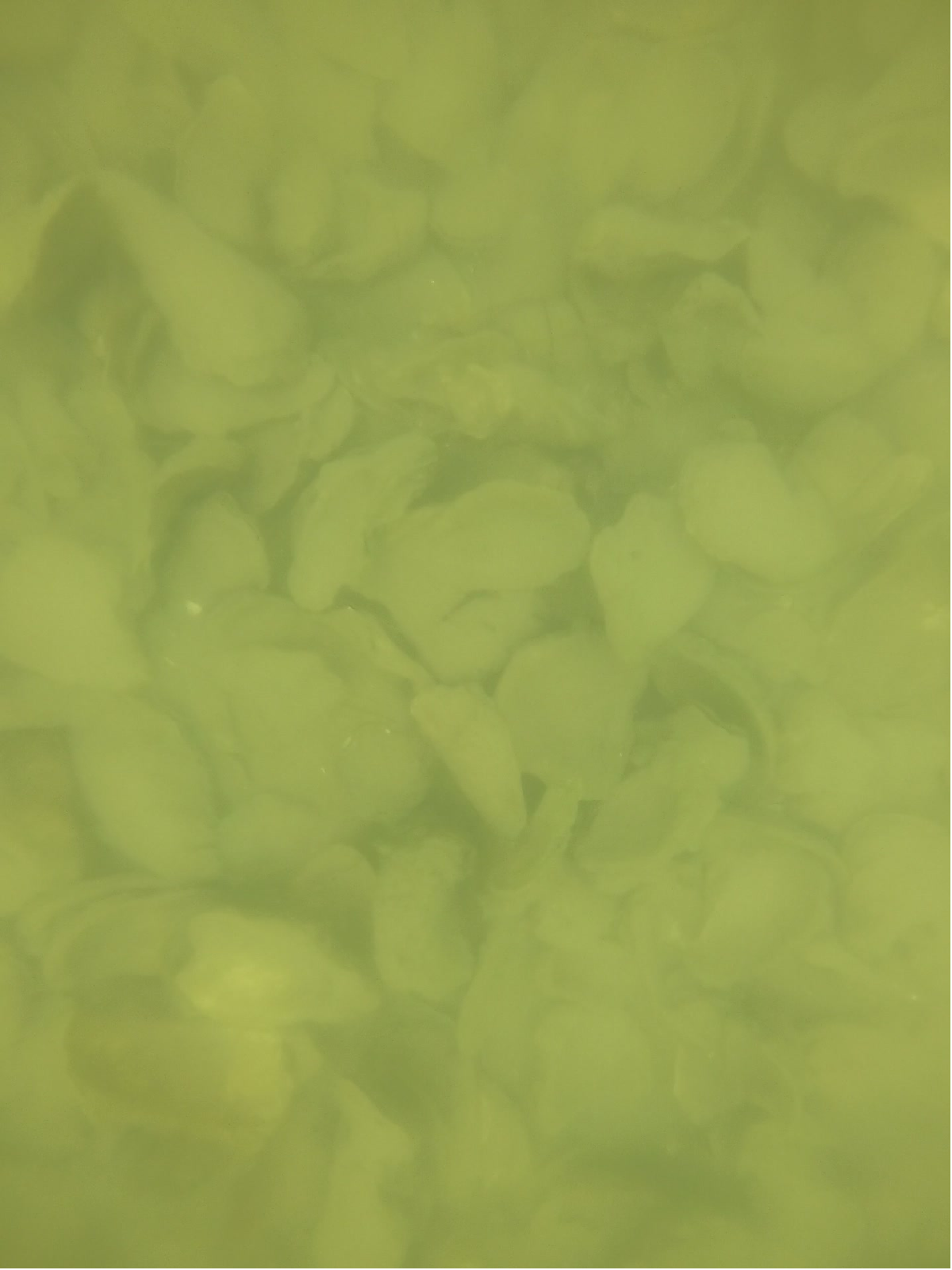}}\hfill
\subfloat[]{\includegraphics[width=.225\textwidth]{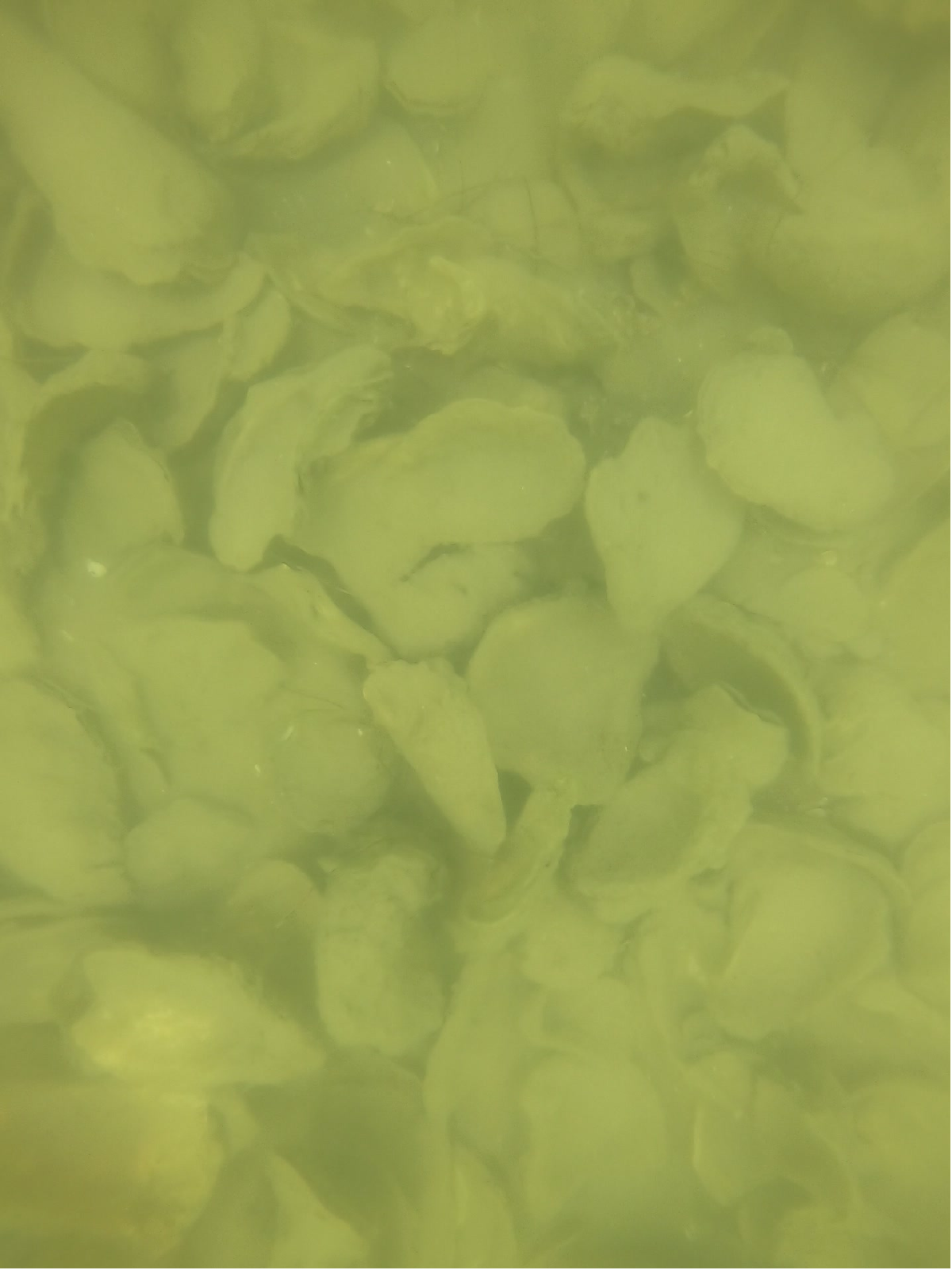}}\hfill
\subfloat[]{\includegraphics[width=.225\textwidth]{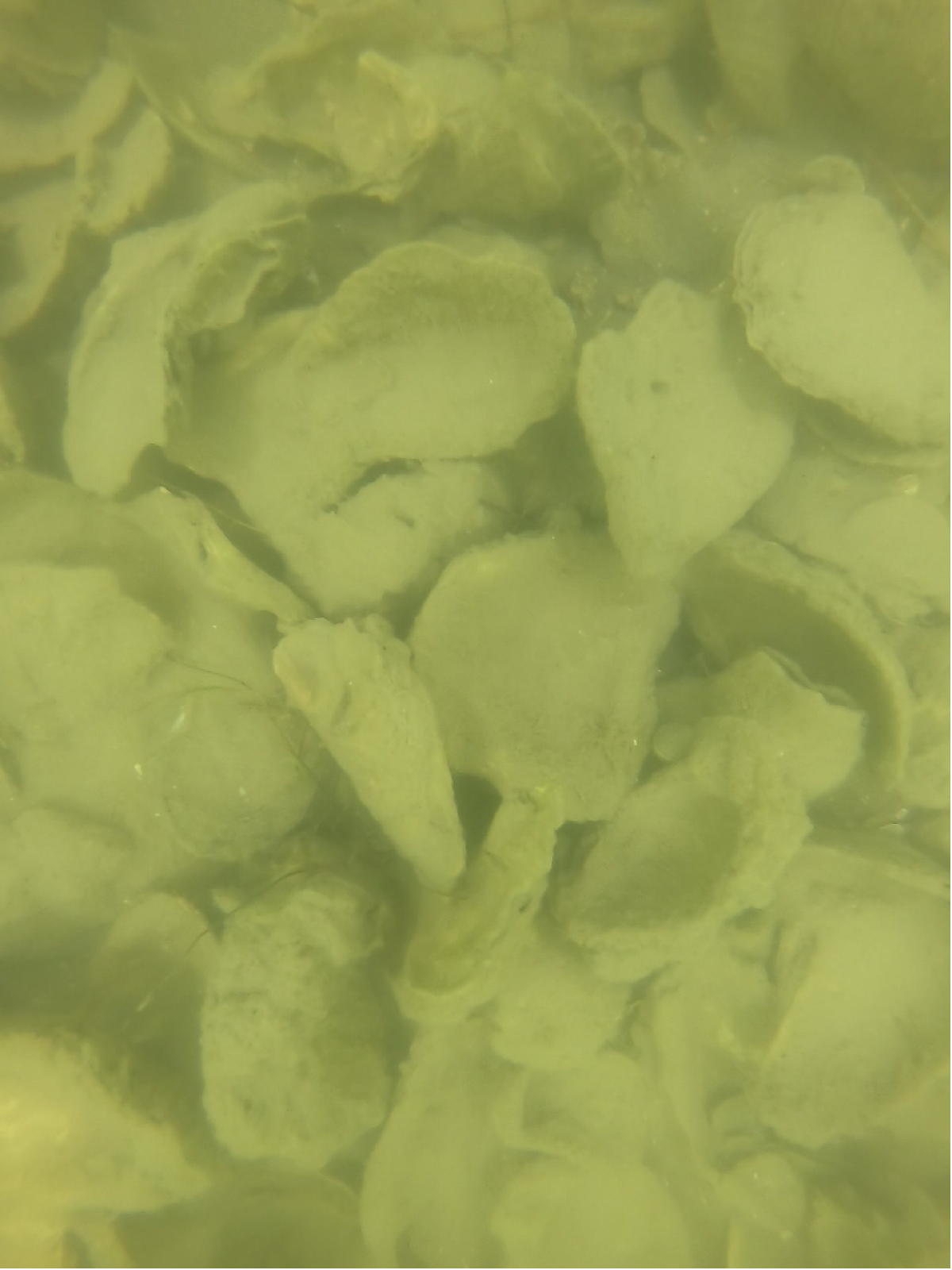}}
\caption[Different Views]{Images of an arbitrary location of the bay bottom but with different altitudes as the ROV moves toward the bottom from (a) to (d). Due to the high turbidity of the water, the bottom is only visible within around three feet.}
\label{fig:turbidity}
\end{figure}

We equipped the ROV with an additional camera, a GoPro HERO7 Black, and led lights to view the seafloor. We set up this camera to store the videos locally and stream to the remote machine using FFmpeg software~\cite{ffmpeg} installed on the local machine. The images taken from the Chesapeake Bay have high turbidity, and the bottom is not visible without external lights. With our lighting setup, we could only see oysters within around three feet from the bottom, and thus we take images within that range. Figure~\ref{fig:turbidity} shows the images of the bay bottom for different heights from the Bay bottom. 
\subsection{Oyster data set and Annotation}\label{subsec:annotation}
\begin{figure} 
\centering
\subfloat[]{\includegraphics[width=.45\textwidth]{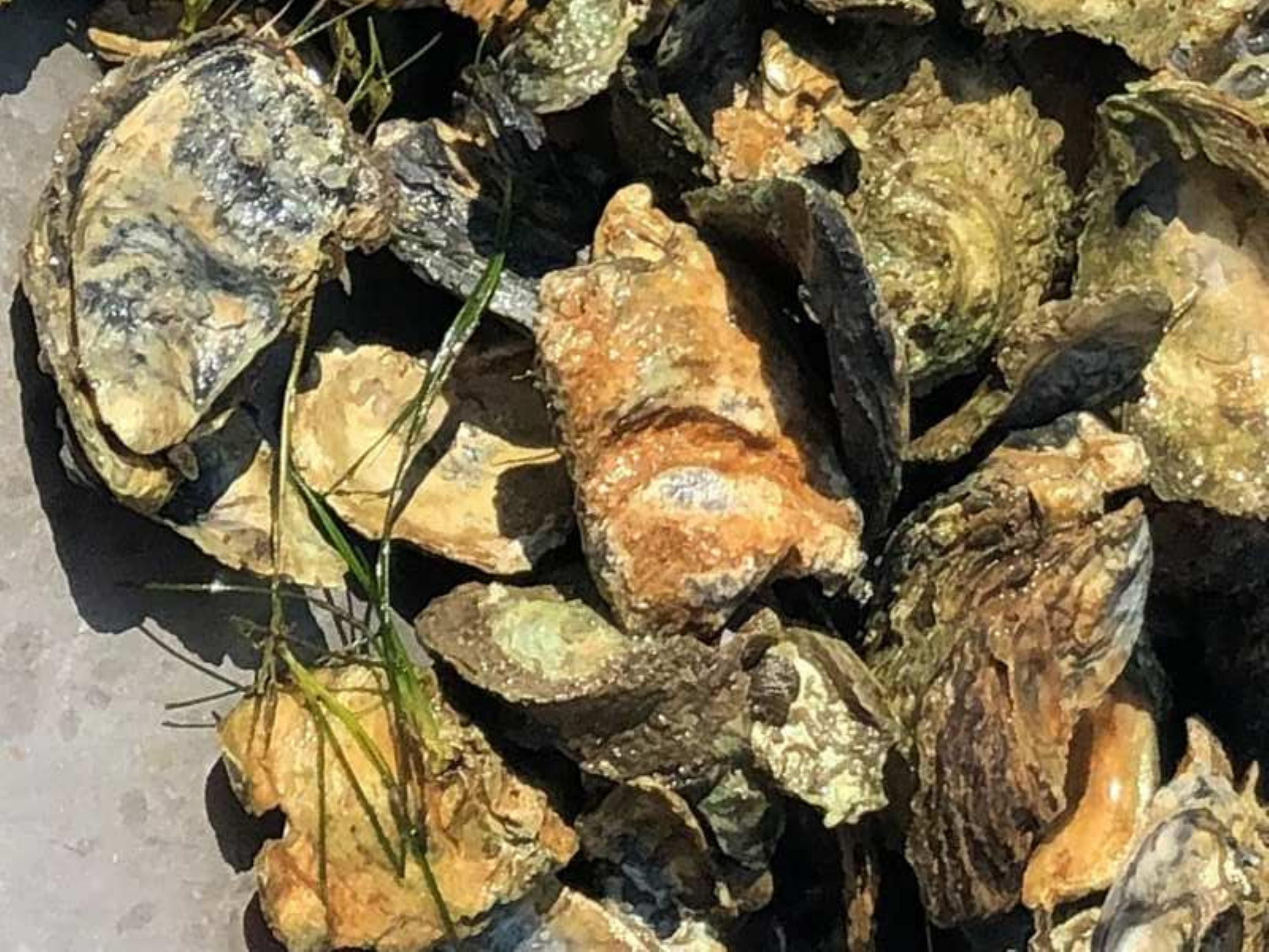}}\hfill
\subfloat[]{\includegraphics[width=.45\textwidth]{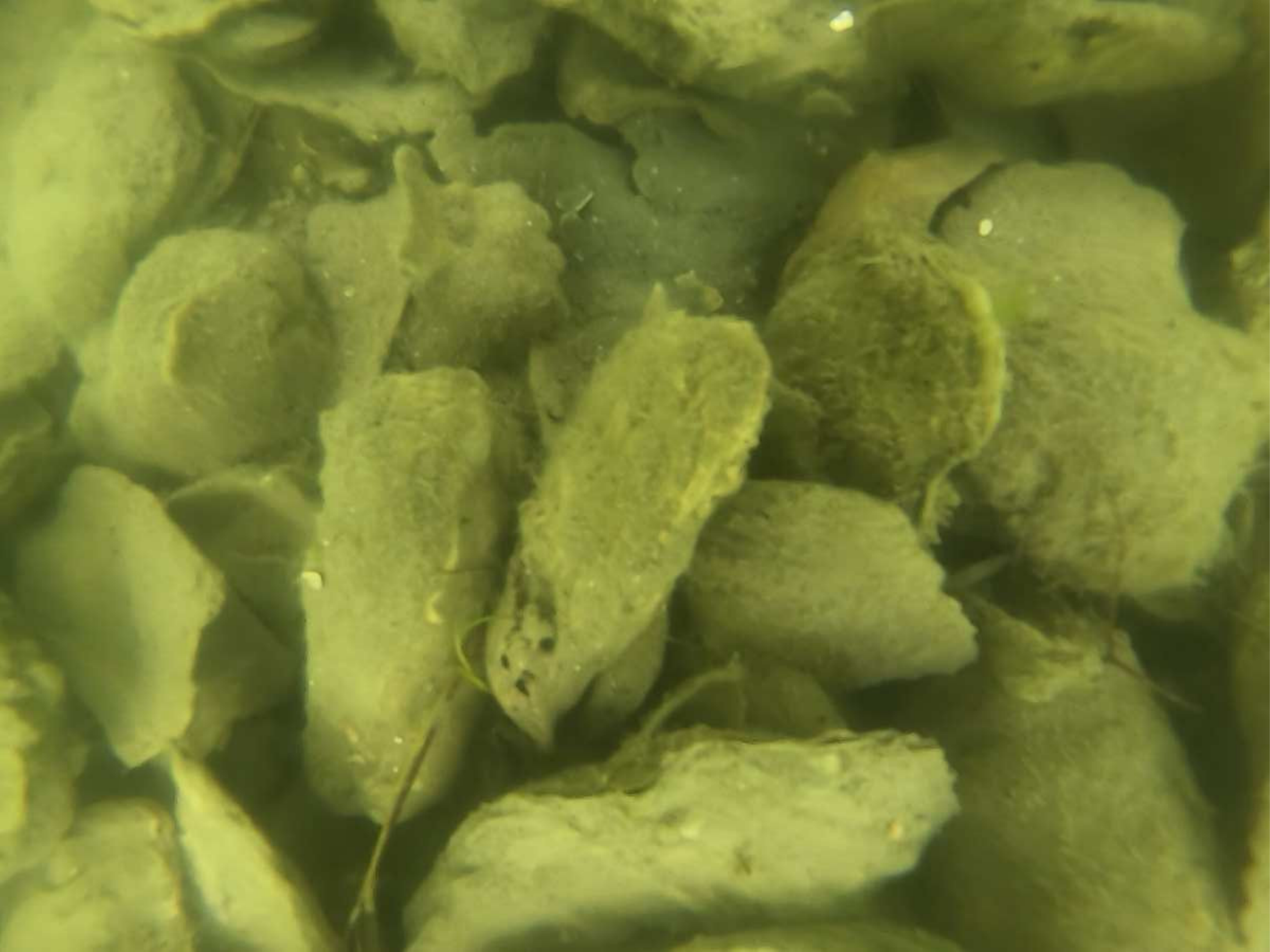}}
\caption[Different Textures]{Sample images of (a) washed oysters and shells after dredging and (b) oysters and shells underwater}
\label{fig:texture}
\end{figure}

Images from the bay bottom recorded videos using our customized ROV were used for constructing the oyster data set. There are two main challenges for ground-truthing the live oysters in the images. 
First, both live oysters and empty shells are lying down in the bay bottom. Live oysters have two valves (or shells) that are slightly open enough to ventilate or feed. Animal orientation is a significant factor for detecting live oysters. If animals are cup-side up, this orientation will mask whether they have two valves or one and how much the gaps are between two valves. Second, the oysters and shells are heavily covered in sediment, and the animal is heavily fouled with epibionts like bryozoa. Therefore, the actual color texture of oysters is not visible (Figure~\ref{fig:texture}). To overcome these challenges (without manipulating animals, which is out of scope of this article), we rely on two things. First, since we have the video data we can use it to analysis an image for ground truthing when necessary. Second, we assume that the object is an live oyster wherever its orientation is cup-side up.

\begin{figure} 
\centering
\begin{floatrow}
\subfloat[Image]{\includegraphics[width=.29\textwidth]{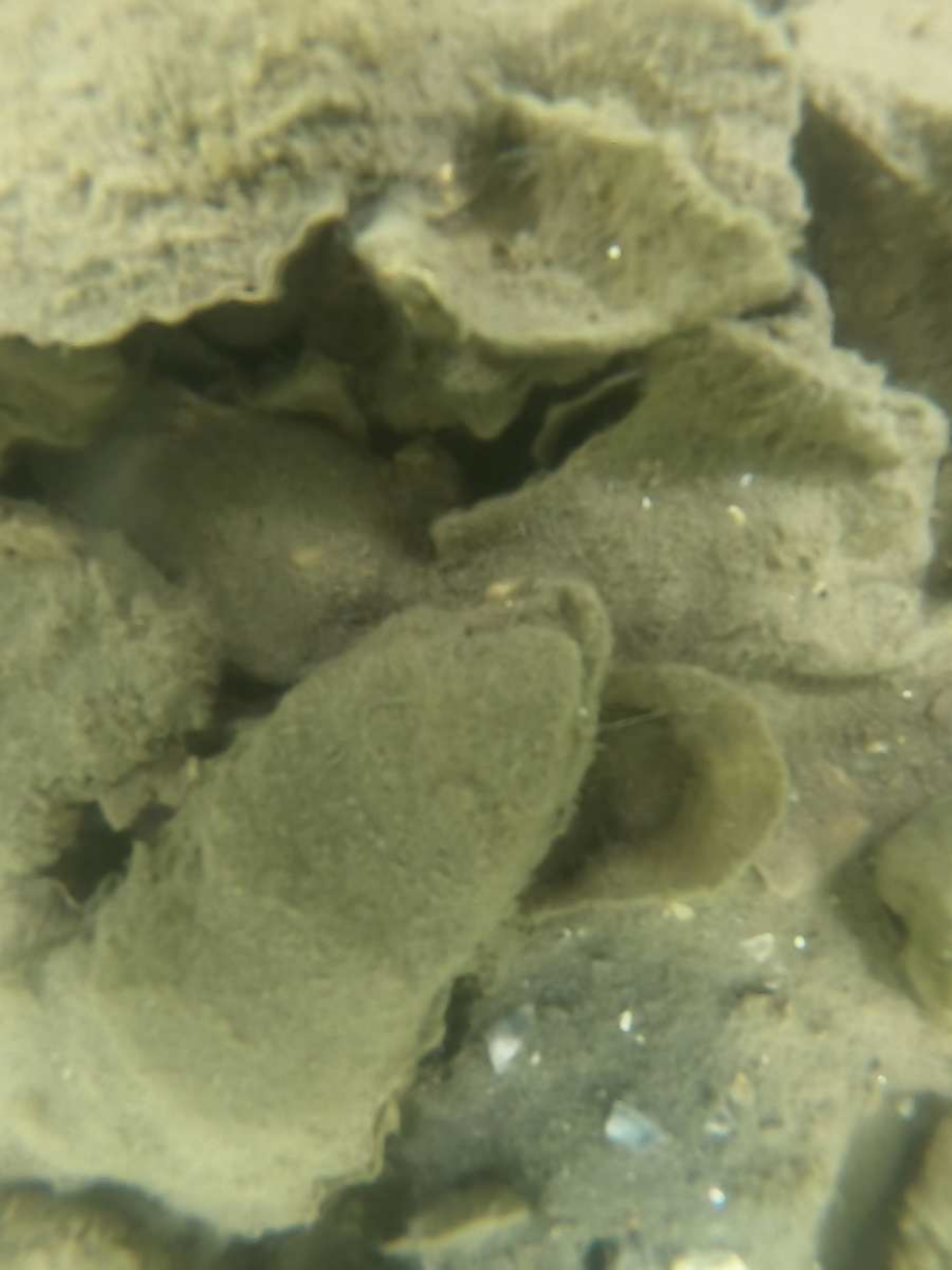}}
\subfloat[Mask]{\includegraphics[width=.29\textwidth]{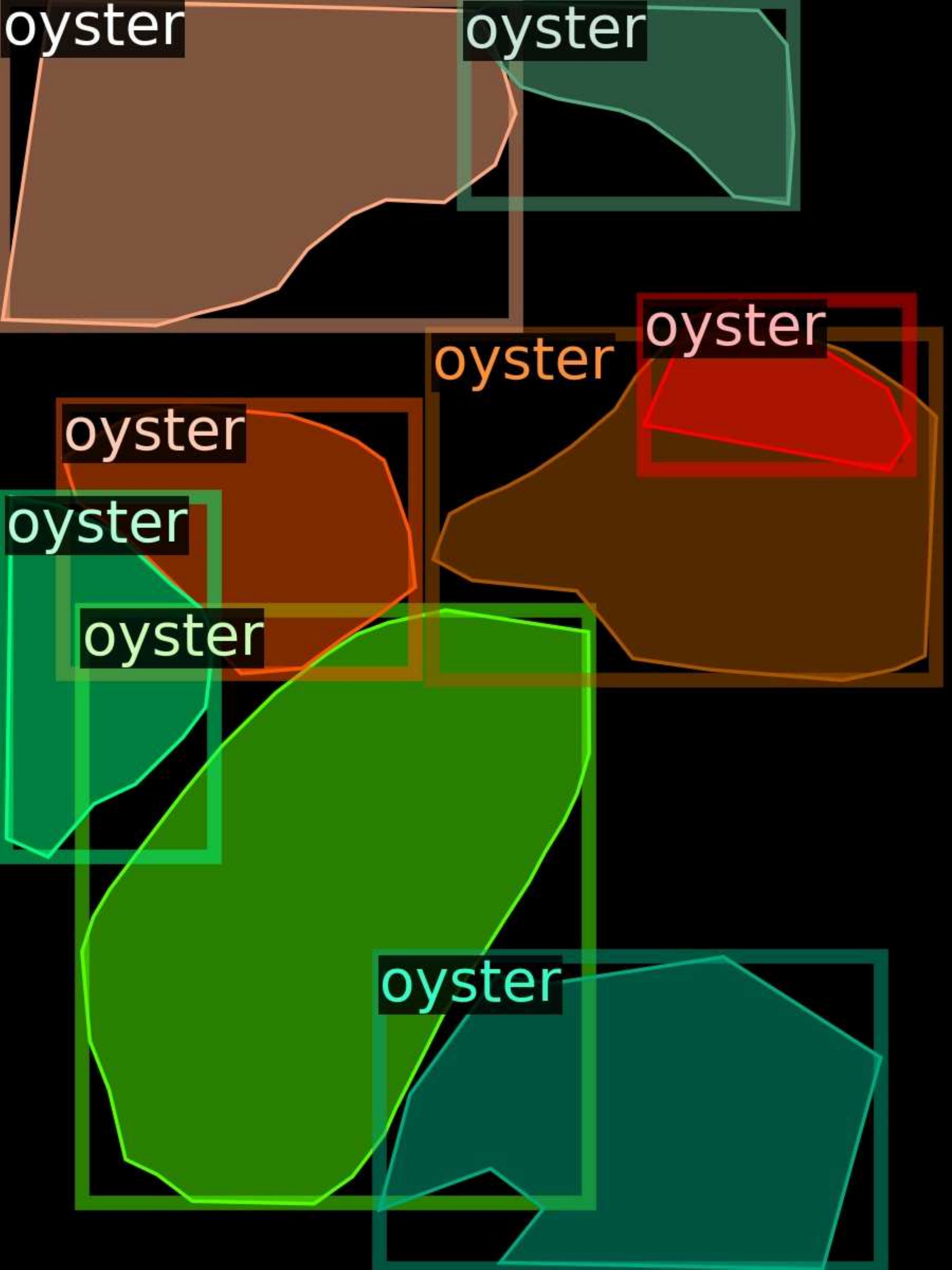}}\quad
\subfloat[Masked image]{\includegraphics[width=.29\textwidth]{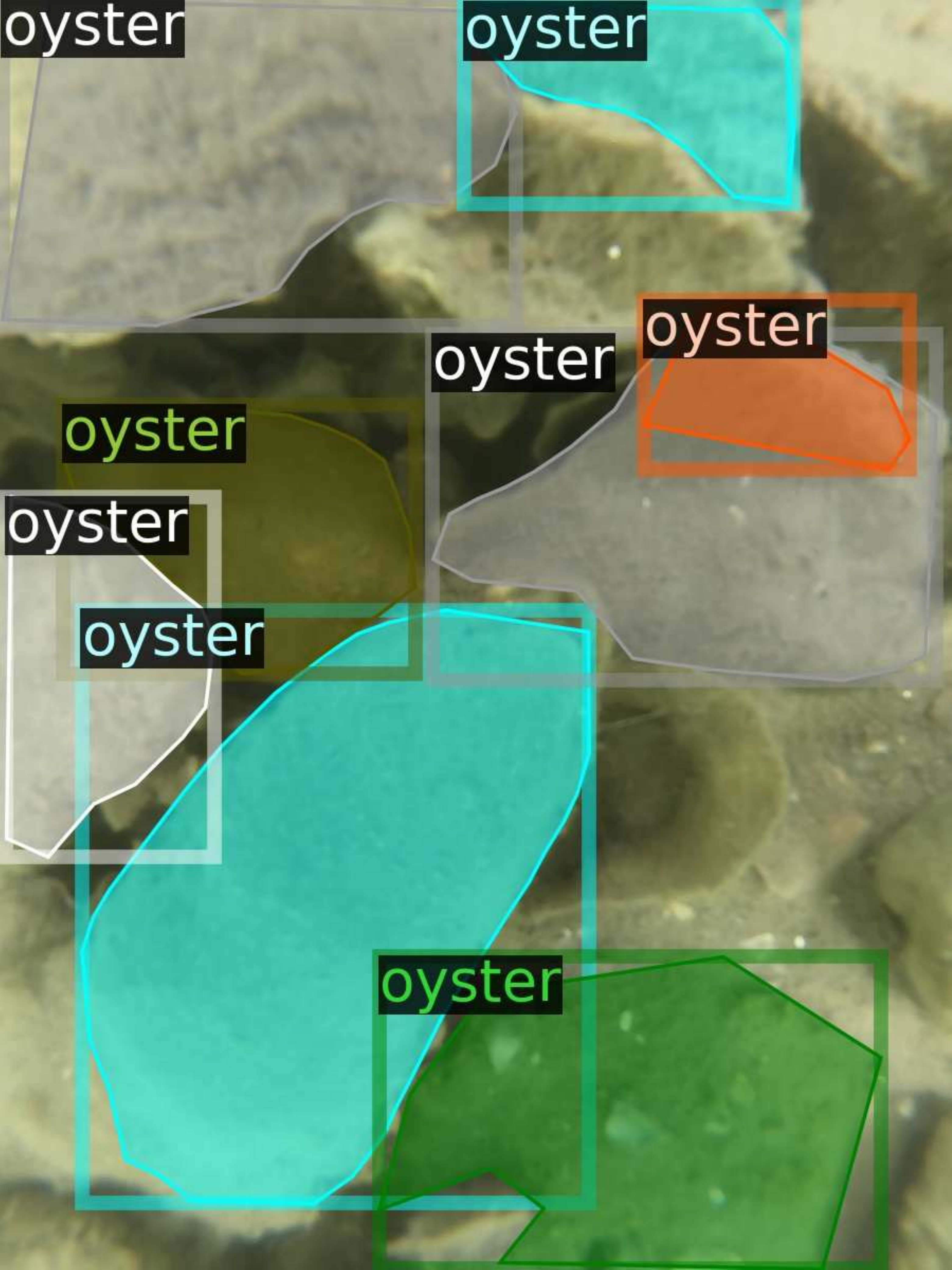}}
\caption[Sample Annotation]{Sample annotation and labeling of an image in the data set}
\label{fig:masksample}
\end{floatrow}
\end{figure}

To make the annotation process more convenient, we annotate the data set in two phases using the image annotation tool labelme~\cite{labelme2016} and another tool we made for converting to masks to labelme software format. First, we selected fifty sample images and generated mask images of the oysters. We used a pre-trained Mask R-CNN (described in Section~\ref{sec:network}) as an initial object detector network and trained it using these fifty images. Next, we selected another set of images and ran them through our initial object detector to get preliminary predictions of oysters. We used our tool for converting the predictions of the trained network to labelme file format. We reviewed and adjusted the generated annotations for making the ground truth mask images of live oysters. We combined all the images and annotated them for model training of the object detection networks. About sixty-five percent of the data set was used for training, and the rest was used for evaluating the performance of the trained networks. Figure~\ref{fig:masksample} shows a sample annotation of live oysters.
\begin{figure}
\centering
\subfloat[Original image\label{fig:o1}]{\begin{sideways}\includegraphics[height=.24\textwidth]{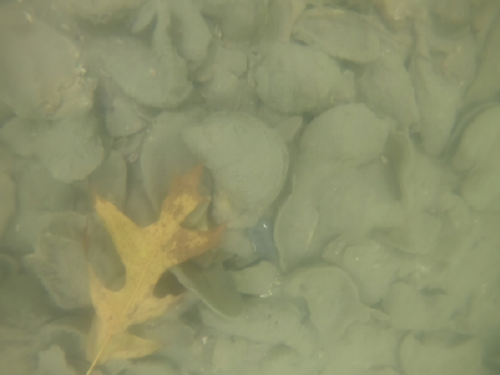}\end{sideways}}\hfill
\subfloat[Inferred image\label{fig:i1}]{\begin{sideways}\includegraphics[height=.24\textwidth]{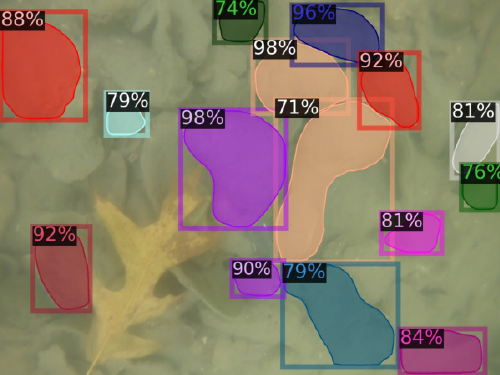}\end{sideways}}\hfill
\subfloat[Original image\label{fig:o2}]{\includegraphics[width=.24\textwidth]{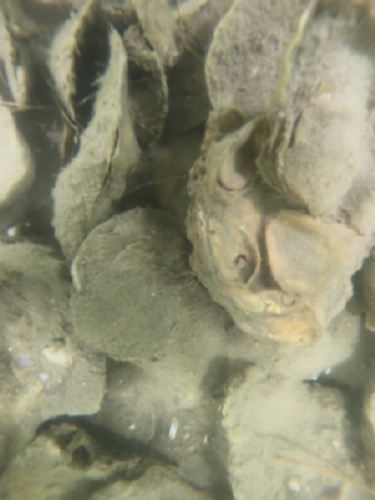}}\hfill
\subfloat[Inferred image\label{fig:i2}]{\includegraphics[width=.24\textwidth]{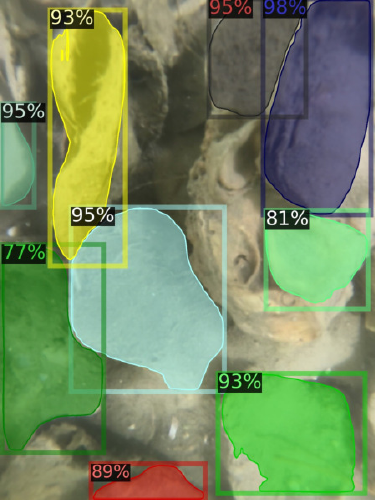}}
 
\subfloat[Original image\label{fig:o3}]{\includegraphics[width=.24\textwidth]{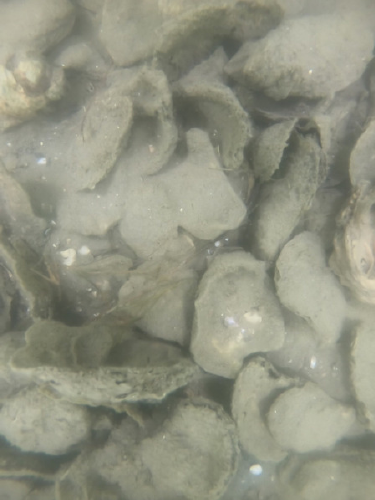}}\hfill
\subfloat[Inferred image\label{fig:i3}]{\includegraphics[width=.24\textwidth]{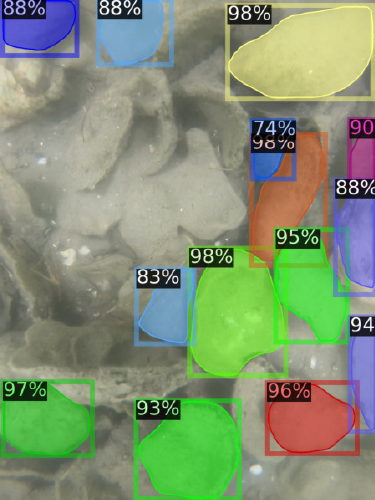}}\hfill
\subfloat[Original image\label{fig:o4}]{\includegraphics[width=.24\textwidth]{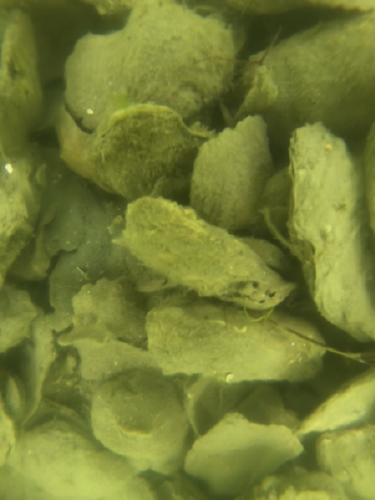}}\hfill
\subfloat[Inferred image\label{fig:i4}]{\includegraphics[width=.24\textwidth]{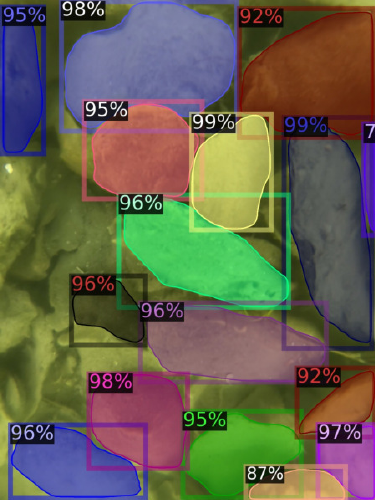}}
\caption[Sample Output]{Sample inferred oysters}
\label{fig:infer}
\end{figure}
\subsection{Implementation and Pre-Trained Models}
Different implementations of Mask R-CNN are already available online. We used Detectron2~\cite{wu2019detectron2} that can be used as a library; trains fast, and has COCO data set~\cite{lin2014microsoft} pre-trained models available. COCO is one of the benchmark data sets for object detection and segmentation and has around 330k images (with more than 200k labeled) of 80 object categories. Since we have a small data set, we leveraged pre-trained models for transfer learning. We used the underlying convolution kernels of the COCO pre-trained networks for our network training, which capture low-level features common in most objects such as edges and angles and only trained for the high-level features special to the oysters. As discussed in Section~\ref{subsec:annotation}, we used 1290 instances of oysters, 827 instances for training, and the rest for validation.
\begin{table}[]
\resizebox{.8\columnwidth}{!}{%
\begin{tabular}{@{}llllll@{}}
\toprule
 & Backbone       & $AP$ & $AP_{50}$ & $AP_{75}$ &  \\ \midrule
 & ResNet50-FPN    & 53.2 & 77.5 & 59.2 &  \\ 
 & ResNet50-DC5   & 55.0 & \textbf{81.8} & 63.3 &  \\
 & ResNet50-C4   & \textbf{55.9} & 80.3 & \textbf{64.1} &  \\ 
 & ResNet101-FPN   & 53.2 & 77.5 & 59.6 &  \\ 
 & ResNet101-DC5  & 53.4 & 78.8 & 60.1 &  \\
 & ResNet101-C4  & 51.6 & 76.8 & 58.7 &  \\ 
 \bottomrule
\end{tabular}
}
\caption{Instance segmentation mask $AP$ values}
\label{tab:AP}
\end{table}
\begin{figure} 
\centering
\subfloat[sample 1]{\includegraphics[width=.315\textwidth]{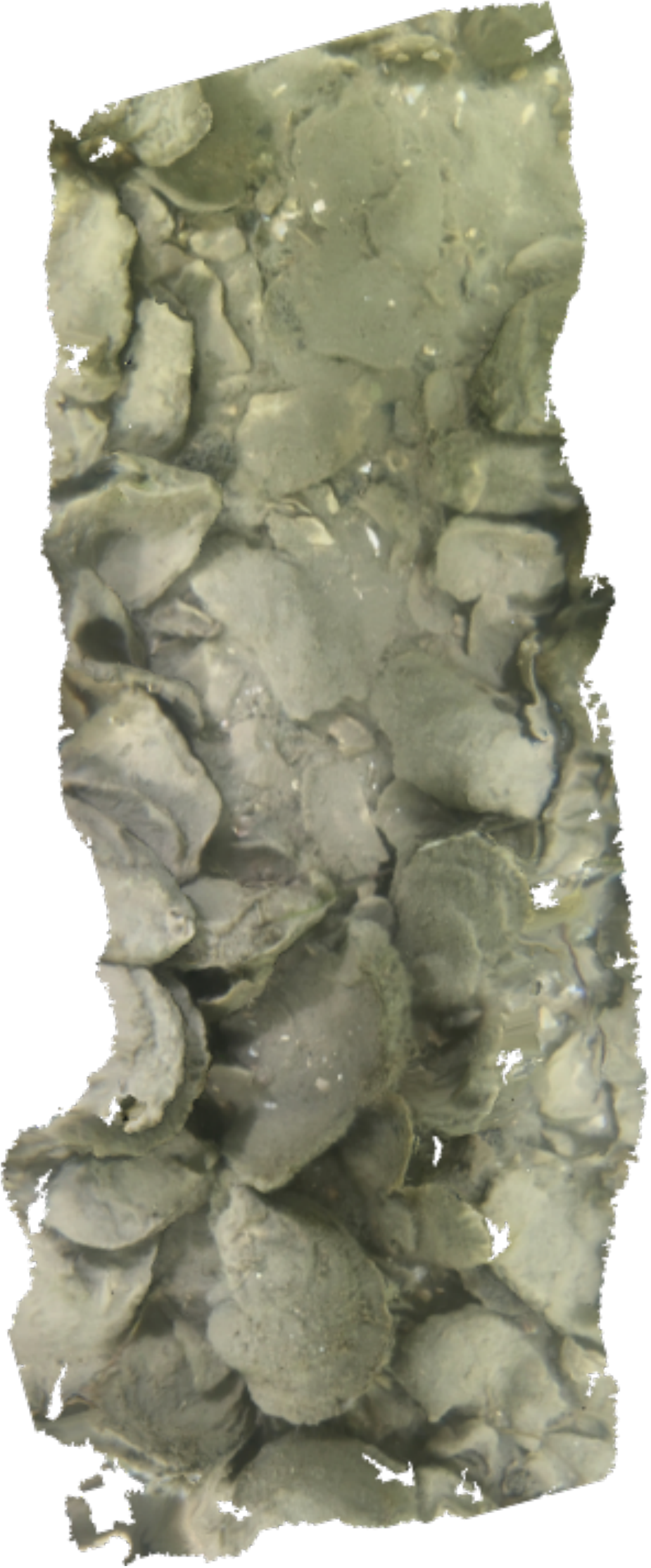}}\hfill
\subfloat[sample 2]{\includegraphics[width=.535\textwidth]{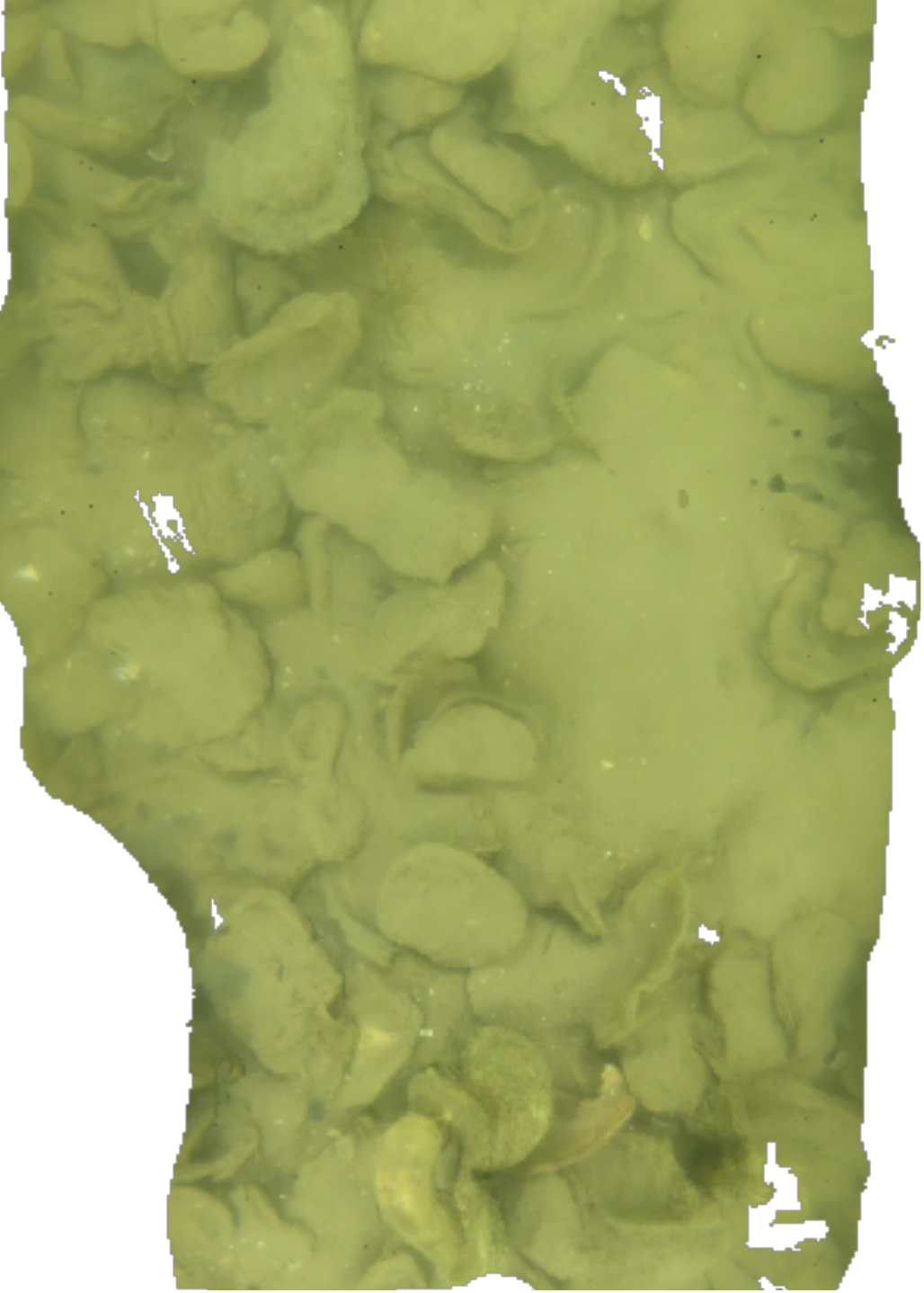}}
\caption[Stitched Videos]{Stitched images of videos (generated via ODM~\cite{odm2020})}
\label{fig:odm}
\end{figure}
\begin{figure*} [th!]
\centering
\subfloat[first frame sample video 1]{\includegraphics[width=.489\textwidth]{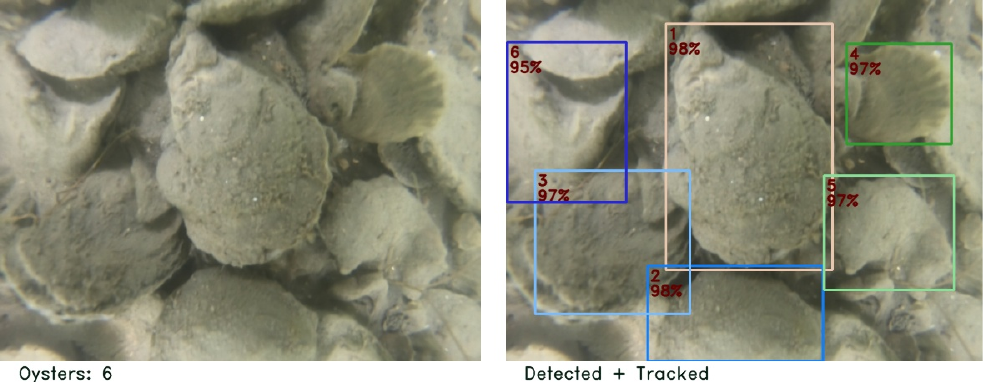}}\hfill
\subfloat[last frame of sample video 1]{\includegraphics[width=.489\textwidth]{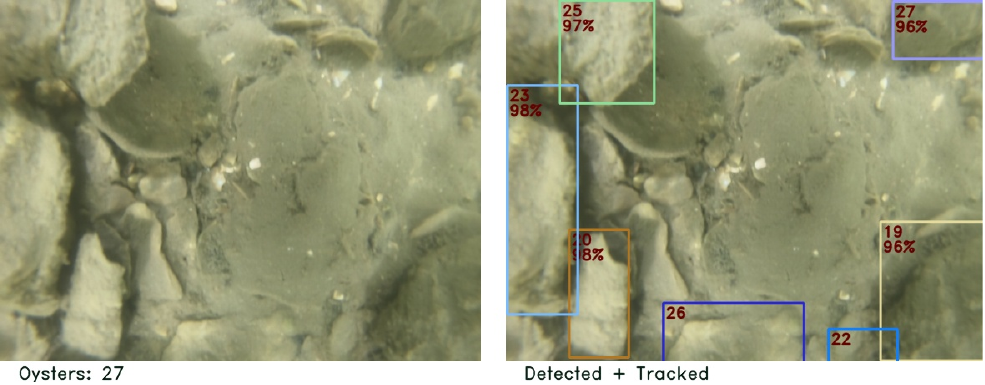}}

\subfloat[first frame sample video 2]{\includegraphics[width=.489\textwidth]{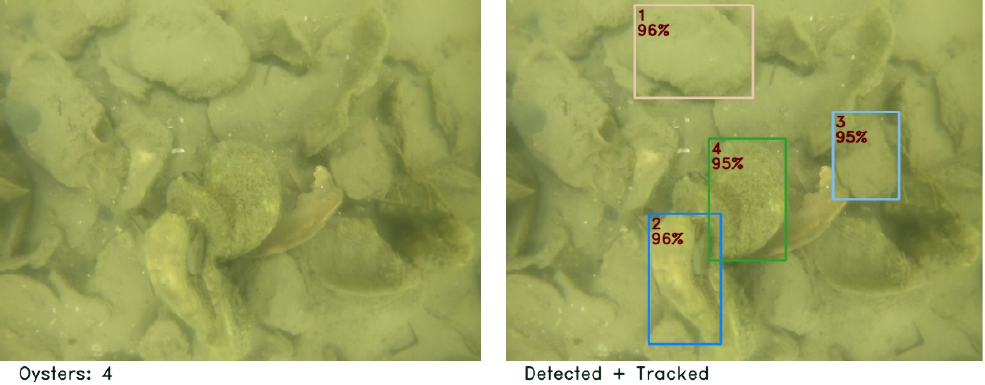}}\hfill
\subfloat[last frame sample video 2]{\includegraphics[width=.489\textwidth]{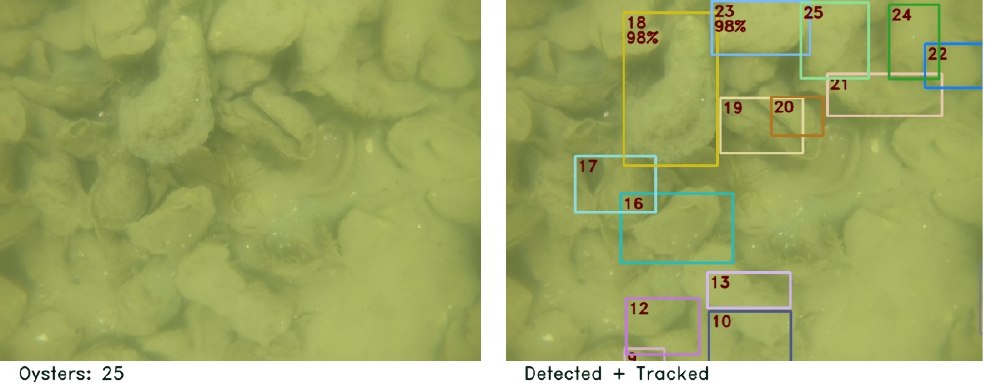}}
\caption[Stitched Videos]{The first and last frames of the sample videos. For each tracker bounding box, the number on the top row shows the tracker identification number. If the detection and tracker matches, the number on the second row shows the oyster inference percentage.}
\label{fig:counting}
\end{figure*}
\subsection{Detection Results}
All the training and evaluation have been executed on a single machine with CPU Intel Xeon E5-2680 2.4 GHz, 2 NVIDIA Tesla P100 GPUs, and 128GB memory. We trained our models for object detection and segmentation with six different backbones of ResNet50-C4, ResNet50-DC5, ResNet50-FPN, ResNet101-FPN, ResNet101-C4, and ResNet101-FPN.
\subsection{Evaluation of the Trained Mask R-CNN}
Average Precision ($AP$) over different IoU (Intersection of Union) thresholds are the common metrics used for object detection and segmentation algorithms~\cite{lin2014microsoft}. 
For a detected object (here oyster), IoU is the ratio of the predicted object's overlap area and the annotated ground truth over their union area. It is used for detecting the number of true positives (TP), false positives (FP), and false negatives (FN). TP is the number of objects detected positive correctly, FP is the number of objects detected incorrectly, and FN is the number of missed objects. For example, if the IoU threshold is set to 50 percent, the predicted object is a true positive if IoU is greater than 0.5; otherwise, it is a false positive. If the object is not predicted, it is a false negative. Precision (P) is the ratio of TP over the number of the predicted objects, and recall is the ratio (R) of the number of TP over the number of ground truth objects. For an IoU threshold, if we plot the precision over recall curve, the area under the curve is the $AP$ for that threshold. In this work, we report mask $AP$ values to evaluate our trained model using the predicted masks.

Table~\ref{tab:AP} compares the different $AP$ values for the oyster segmentation of our trained models for the validation set. $AP_{50}$ and $AP_{75}$ are the average precision calculated for IoU thresholds of $0.50$ and $0.75$, respectively. $AP$ is the mean of $\{AP_{50}, \{AP_{55} \cdots AP_{95}\}$. The best $AP$ and $AP_{75}$ were achieved for the network with ResNet50-C4 backbone with the values of $55.9$ and $64.1$, respectively. The best $AP_{50}$ was achieved for ResNet50-DC5 backbone with a value of $81.8$. As mentioned in~\ref{subsec:annotation}, there two challenges for detecting live oysters, (a) animal orientation and (b) sediments on oysters. These two challenge heavily impact the performance of the Mask R-CNN network. For annotating the ground truth, we used the video sequence to inspect the previous and next images when it was challenging to distinguish live oysters from empty shells. If it is challenging for the annotator to distinguish between the live oysters and the reefs on the bottom, the same problem applies to the network which learns from those annotations. Nonetheless, looking at the inferred images of the oysters, the results seem promising. Our ultimate goal is to make an underwater robot capable of picking up (harvesting) oysters. Therefore, as long as the network can detect most of the live oysters, we can overlook the false positives.
Figure~\ref{fig:infer} shows samples of inferred oysters. Figure~\ref{fig:i1} shows instance segmentation of Fig.~\ref{fig:o1}. This image is taken from a relatively far distance, and due to turbidity, it is relatively more blurry. There are more false positives in the inferred image compared to the other images in Fig.~\ref{fig:infer}. However, the number of missed detected oysters (false negatives) is small (two oysters in the top left). Figure~\ref{fig:i2} showcases a moderately high performance of the network. There is no false positive in this image, but there is one false negative (the oyster beneath the oysters on the right). Figure~\ref{fig:i3} demonstrates a moderately average performance of the network. There are three false positives in this image (from the center to the right), and there is no false negative. Figure~\ref{fig:i3} illustrates a moderately high performance of the network. There is no false positive in this image but there one false negative (the oyster beneath the oysters on the right). Figure~\ref{fig:i4} depicts a relatively high performance of the network. There are three false positives in this image (right bottom), and there is no false negative.
\subsection{Tracking and Counting Implementation}
We used OpenCV~\cite{opencv} for implementing KCF tracker. We picked two sample videos to demonstrate our oyster counting method. To better demonstrate the results, we picked two short videos with 1000 and 300 frames (around 16 and 5 seconds for 59 fps) and used OpenDroneMap (ODM)~\cite{odm2020} to extract a single stitched image for each of these videos. Note that generating such images takes a much longer time than counting the oyster using our method. Figure~\ref{fig:odm} shows the stitched images. Figure~\ref{fig:counting} shows the first and last frames of the sample videos as well as the results of the proposed method. We count the oysters in a sample video ourselves manually and compare them with the results from implementing our method, as shown in Table~\ref{tab:count}. We also repeat these for another two sample video sequences without generating their stitched image. The results show an average accuracy of $79.8\%$. There are three main sources of error: (i) miss-classification of empty shells as oysters, (ii) recounting of oysters, which was reduced by tracking the unmatched tracking bounding boxes, and (iii) the oyster instances that the detection network missed.
\begin{table}[]
\resizebox{.9\columnwidth}{!}{%
\begin{tabular}{@{}llcc@{}}
\toprule
 &        & Tracking-by-detection & Manual counting \\ \midrule
 & sample 1    & 27 & 21  \\ 
 & sample 2   & 25 & 20 \\
 & sample 3   & 118 & 94 \\
 & sample 4   & 215 & 176 \\
 \bottomrule
\end{tabular}
}
\caption{Number of oysters in the sample videos}
\label{tab:count}
\end{table}
\section{CONCLUSION}\label{sec:conclusion}
Oysters are an essential species in marine ecosystems, and restoration of oyster habitats is vital. Our goal is to help to monitor oysters. In this work, we demonstrated how we customized a drone-like ROV, BlueRobotics BlueROV2, with a camera, GoPro HERO7 Black, for videography of the Chesapeake Bay bottom. Moreover, we annotated the visual data captured from the bay bottom to make an oyster data set. We chose the state-of-the-art CNN for target tracking and instance segmentation, Mask R-CNN, and employed detectron2 for training and evaluation of the oyster data set. We compared the results for different backbones in the network structure. Our trained networks' prediction results showed an Average Precision ($AP_{50}$) of 81.8 for our oyster data set. We also combined the detection CNN with an object tracker for counting the number of oysters in videos.
Target detection and instance segmentation of oysters is the first step towards making a robotic tool which environment-friendly and can harvest oysters more efficiently. The code is available online at \href{https://github.com/bsadr/oyster-detection}{GitHub/bsadr}~\cite{oysterd}.
\section*{ACKNOWLEDGMENT}
The authors wish to thank Mr. Bobby Leonard for providing his boat and giving us tours in his oyster lease for taking videos in the Chesapeake Bay. This work was supported in part by the Maryland Robotics Center postdoctoral fellowship program. This work was supported in part by USDA NIFA sustainable agriculture system program under award number 20206801231805.
\bibliographystyle{plain}
\bibliography{related}
\end{document}